\documentclass[runningheads]{llncs}
\usepackage[T1]{fontenc}
\usepackage{graphicx}
\usepackage{xcolor,mdframed}
\usepackage{amsfonts}
\usepackage{amsmath}

\usepackage{ifthen}
\usepackage[export]{adjustbox}
\usepackage{subcaption,graphicx}

\usepackage{url}

\begin{document}
\title{Evaluation of Confidence-based Ensembling in Deep Learning Image Classification}
\titlerunning{Confidence-based Ensembling in DL image classification}
\author{Rafael Rosales\inst{1}\orcidID{0000-0002-3917-8776} \and
Peter Popov\inst{2}\orcidID{0000-0002-3434-5272} \and
Michael Paulitsch\inst{1}\orcidID{0000-0002-9241-5806}} 
\authorrunning{R. Rosales et al.}
\institute{Intel Labs, Neubiberg, Germany\\
\email{\{rafael.rosales, michael.paulitsch\}@intel.com}\\
\and
City, University of London, London, UK\\
\email{P.T.Popov@city.ac.uk}\\
}
\maketitle              
\begin{abstract}

Ensembling is a successful technique to improve the performance of machine learning (ML) models.

\textit{Conf-Ensemble} is an adaptation to Boosting to create ensembles based on model confidence instead of model errors to better classify difficult edge-cases. 
The key idea is to create successive model experts for samples that were difficult (not necessarily incorrectly classified) by the preceding model.
This technique has been shown to provide better results than boosting in binary-classification with a small feature space (\~80 features).

In this paper, we evaluate the Conf-Ensemble approach in the much more complex task of image classification with the ImageNet dataset (224x224x3 features with 1000 classes).
Image classification is an important benchmark for AI-based perception and thus it helps to assess if this method can be used in safety-critical applications using ML ensembles.

Our experiments indicate that in a complex multi-label classification task, the expected benefit of specialization on complex input samples cannot be achieved with a small sample set, i.e., a good classifier seems to rely on very complex feature analysis that cannot be well trained on just a limited subset of "difficult samples".

We propose an improvement to Conf-Ensemble to increase the number of samples fed to successive ensemble members, and a three-member Conf-Ensemble using this improvement was able to surpass a single model in accuracy, although the amount is not significant.
Our findings shed light on the limits of the approach and the non-triviality of harnessing big data.

\end{abstract}
\section{Introduction and state-of-the-art}

State-of-the-art deep learning models are already able to, on average, perform multiple different tasks better than a human would.
One of the main challenges today, is that such models are not able to cope well with edge cases that have not been well represented in their training dataset.

One of the most successful approaches to mitigating this problem has been the use of deep learning \textit{ensembles}.
A deep learning ensemble is a collection of \textit{different} machine learning model instances (\textit{weak learners}), a.k.a. ensemble members or channels, that have been trained to perform the same task, e.g. classification, using the same inputs.
An ensemble of machine learning models can potentially produce more accurate results than a single model due to the diversity of solutions found to the same problem as this model diversity  results in diversity in failure behavior.

The diversity of ensemble members is frequently achieved implicitly through the randomness of the training process.
Explicit approaches have also been proposed to decrease the bias or the variance of the ensemble members by applying different sampling heuristics of the training data.

Bagging~\cite{breiman1996bagging} is a technique to partition the training data into different subsets and train each ensemble member with one of those subsets.
The result is a reduction of the ensemble bias through the averaging of the increased variance of the predictions.

Boosting~\cite{schapire1999brief}, is a technique where a model is trained first with the complete training data set, and samples where the model has made incorrect predictions are used more often in the training of the next ensemble member.
The result is a reduction of the ensemble bias through the re-weighting of the sample importance.

Negative Correlation Learning~\cite{LIU}, is a technique where all of the models of an ensemble are trained in parallel not independently.
The loss function of each model is extended to include a  term that penalizes the similarity of the model predictions, thus explicitly inducing variance.
The result is an ensemble of models with diverse failure behavior, i.e., not all models will fail the same on the same set of inputs.

In~\cite{KunchevaW03}, a study of the impact of prediction diversity is in ensembles is presented.
\cite{RMP23} provides an evaluation on more recent deep learning (DL) models such as transformers, automatic architectures created with neural architecture search (NAS) and a proposed new diversity metric based on attribution, i.e., on the arguments about the input of why a model made a given prediction.

Conf-Ensemble~\cite{confEns}, is a recently proposed technique to use the uncertainty information to create models with a complementary expertise to produce more accurate ensembles.
The approach has proven successful results in binary classification of an Intrusion Detection System (IDS).
Each of the samples in this dataset has about 80 features to classify a network session as normal (i.e. "benign") or not (i.e. "an attack").

The work at hand evaluates the Conf-Ensemble approach in a much more complex computer vision multi-class classification task.
Classification is at the core of many safety-critical perception tasks, such as object detection in autonomous driving.
One of the main tasks of an object detector is to be able to identify an object based on image features.
Such feature detectors, a.k.a. backbones, are mostly created as image classifiers.

If Conf-Ensemble can be proven to help the performance of vision tasks, such as image classification, then the use of Conf-Ensemble in safety-critical systems with vision-based perception would provide a more robust approach for difficult input images than a typical machine learning model would predict with low confidence.

The rest of the paper is structured as follows: first we summarize the Conf-Ensemble approach in Section~\ref{sec:approach}.
The experiments performed in image classification are shown in Section~\ref{sec:exp}.
Finally a discussion of the results, conclusions and directions for further research are given in Section~\ref{sec:discConc}.

\section{Conf-Ensemble}
\label{sec:approach}

In the rest of this section, we provide an overview of the Conf-Ensemble approach, followed by the confidence metric used, the ensemble creation flow and the inference process.
Lastly, we propose an adaptation of Conf-Ensemble to address an observed problem in this image classification task.

    \subsection{Approach}

    The general Conf-Ensemble approach is described in two phases: a) how to create an ensemble and b) how the ensemble is used during runtime.
    A classification confidence metric described next is used  for different purposes on each phase.

    \subsection{Classification confidence metric}
        
        The classification confidence metric, and other formulations defined in~\cite{confEns} are presented here for multi-label classification.

        The raw output of a single classifier model, represented as $O_{C}$ on a single input sample $x$ is defined is:
        
        \begin{equation}
            O_{C}: \mathbb{R}^M \rightarrow \mathbb{R}^N;
            \mathbf{x} \mapsto \mathbf{y}
        \end{equation}

        where $M$ is the number of features of the input samples and $N$ is the total number of classification labels.
        
        The raw outputs of a classifier, sometimes called logits, can be in theory any real value from minus infinity to infinity.
        To obtain \textit{probability scores}, i.e., an output vector whose elements sum-up to 1, i.e., a mapping from $(-\infty,\infty)$ to $[0,1]$, the \textit{softmax} function is often applied.
        The probability score produced by a single classifier model using the softmax function, represented as $P_{C}$ on a single input sample $t$ is defined is:

        \begin{equation}
            P_{C}: \mathbb{R}^M \rightarrow \mathbb{R}^N;
            \mathbf{x} \mapsto \mathbf{y}
            \mid \, 0 \le y_i \le 1 \quad \forall i \in \{1,2,\dots,N\}
        \end{equation}

        The distances of the prediction with maximum probability score to the edges of the range [0,1] are defined as:

        \begin{equation}
            D_{0}: \mathbb{R} \rightarrow \mathbb{R};
            \max\{\mathbf{y}\} \mapsto \max\{\mathbf{y}\}
        \end{equation}

        \begin{equation}
            D_{1}: \mathbb{R} \rightarrow \mathbb{R};
            \max\{\mathbf{y}\} \mapsto 1 - \max\{\mathbf{y}\}
        \end{equation}

        In other words, $D_0$ is the distance of the predicted class to a probability score of 0, and $D_1$ is the distance of the predicted class to a probability score of 1.

        The uncertainty score (referred to as confidence score in~\cite{confEns}), is defined as:

        \begin{equation}
            U(P_{C}(x)) = \min(D_{0},D_{1}) \equiv \min(\max\{\mathbf{y}\},1 - \max\{\mathbf{y}\})
        \end{equation}

        $U$ is thus the shortest distance to either a probability score of 0 or 1.
        A $U$ metric value of 0.5 corresponds to the maximum unconfidence, as the prediction probability score of the model is farthest away from 0 and from 1.
        A $U$ metric value close to 0 corresponds to either a probability score very close to 1 (highly confident of a correct prediction) or very close to 0 (highly confident of an incorrect prediction.
        
        \subsection{Ensemble creation process}

        The creation process is sequential.
        The first member of the ensemble is trained with the complete training dataset $I_{M_0}= \{ x \in X$ \}.
        In this work we refer to this ensemble member as $M_0$.
        
        The trained $M_0$ model is then used to compute the uncertainty score $U$ on each of the training samples $x \in X$.
        If the uncertainty score $U(P_{M_0}(x))$ is above a given \textit{training uncertainty threshold} $T_{t}^{L=1}$, then this sample is chosen to be used for training for the next ensemble member $M_1$.
        In other words, all samples in which $M_0$ was not confident enough form the training dataset for $M_1$: 
        
        \begin{equation*}            
            I_{M_1}=\{x \in I_{M_0} \mid U(P_{M_0}(x)) > T_{t}^{L=1}\}
        \end{equation*}

        \begin{figure}
	   \includegraphics[width=\linewidth]{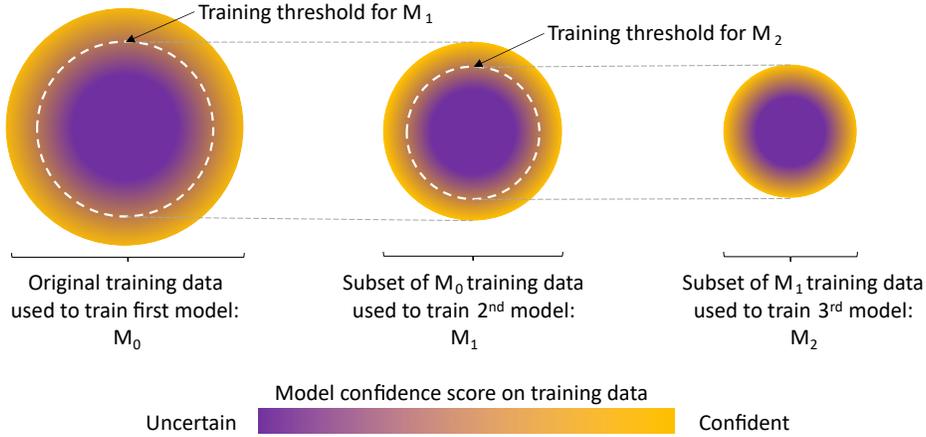}%
          
	   \caption{Training data extraction process for each ensemble member.
        The confidence/uncertainty score is  computed for each training data sample and the set of uncertain samples form the training data set of the next ensemble member.}
        \label{fig:trainIlus}
	\end{figure}

        This is then repeated iteratively for each ensemble member as illustrated in Figure~\ref{fig:trainIlus}.
        The trained model $M_1$ is used to compute the $U$ metric in the samples in this dataset, and each sample with a score above a training uncertainty threshold $T_{t}^{L=2}$ forms the training dataset for the next ensemble member $M_2$.
        This process is performed $S-1$ times for an ensemble with $S$ members using $S-1$ training thresholds.
        The general rule for the creation of the training data subsets for the ensemble members with index bigger than 0 is thus:

        \begin{equation}
            \label{eq:trainingset}
            \forall s \in S \setminus\{s=0\}: I_{M_s}=\{x \in I_{M_{s-1}} \mid U(P_{M_{s-1}}(x)) > T_{t}^{L=s}\} 
        \end{equation}

        \subsection{Ensemble inference process (classification)}

        During inference, the process is sequential as well.
        The input is first fed to $M_0$.
        If the uncertainty score is below a \textit{run-time uncertainty threshold} $T_{r}^{L=0}$ then the prediction of $P_{M0}$ is accepted as the ensemble result.
        If however the score is above that threshold, the result is considered with low confidence and the next ensemble member is run on the given input.
        The same heuristic is applied to accept or reject the prediction of $M_1$, $M_2$, or $M_S$ using $S$ run-time thresholds.
        Note that the run-time thresholds may differ from the training thresholds.
        The former can be dynamically adapted during inference while the later determines the specialization of the models during training and their values need not to be the same.
        
        If none of the predictions where confident, then a consensus \textit{heuristic} is used to choose a prediction.
        The \textit{last member} heuristics accept the prediction of the $M_S$ model.
        The \textit{most confident} heuristic accepts the prediction of the model with the lowest uncertainty score $U$.
        
        \subsection{Proposed extension}
        \label{sec:propExt}

        To increase the size of the training dataset of the subsequent ensemble members, we propose a slight modification of the Conf-Ensemble approach.

        Instead of only evaluating the uncertainty score on the training dataset $I_{M_{s-1}}$ of an ensemble member $M_{s-1}$ to obtain the training dataset for the ensemble member $M_{s}$ as specified in Equation~\ref{eq:trainingset}, we propose to evaluate the complete training dataset.

        \begin{equation}
            \label{eq:proptrainingset}
            \forall s \in S \setminus\{s=0\}: I_{M_s}=\{x \in I_{M_0} \mid U(P_{M_{s-1}}(x)) > T_{t}^{L=s}\} 
        \end{equation}

        With this change, the training dataset sizes will not be continuously reduced, avoiding an under-fitting of the ensemble models.
        This extension is a compromise on the Conf-Ensemble principle to create channels trained on samples which the previous members of the ensemble found difficult to  classify with high confidence. The change, however, is necessary as a measure of making the size of the training dataset  adequately large for the complex tasks of image classification.

\section{Experimental Evaluation}
\label{sec:exp}

    In this section we present the design of the experiments followed by the obtained results.

    \subsection{Design}
    \label{sec:design}

        \paragraph{Task} The task in this case study is image classification using the ImageNet dataset~\cite{DengDSLL009}.
        The number of classification labels is 1000.
        This means that $N=1000$.
        The number of training samples is 1.2 million images and 50 thousand validation images.
        
        \paragraph{Model} We select ResNet50 as the architecture of every single model in an ensemble.
        It accepts color images (RGB) of size $224x224$ pixels.
        This means that the input feature space is $M=150528$ features.
        We create three different two-member ensembles and one three-member ensemble.
        
        \paragraph{Optimization} We use the AdaBelief Optimizer with the following hyperparameters:
        Learning rate= 1e-3, eps=1e-8, betas=(0.9,0.999), weight\_decay=1e-2.
        The scheduler decreases the learning rate by a gamma factor of 0.3 every 15 epochs.
        The loss function is the cross-entropy loss.
        
        \paragraph{Training confidence thresholds $T_{t}$}

        To produce three different ensembles of only two-members named \{e1, e2 \& e3\} we evaluated the following three training thresholds : $T_{t}^{L=1} = \{0.2, 0.1, 0.01\}$.
        A low value of the training threshold means that a lot of samples will be used to train the next model.
        For the creation of a single three-member ensemble we selected the two-member ensemble with threshold $T_{t}^{L=1}=0.01$  and we chose the training threshold for the third member to be $T_{t}^{L=2} = 0.01$.

        \paragraph{Runtime confidence thresholds $T_{r}$}

        We evaluated four run-time thresholds on all ensemble members homogeneously: $T_{r}=\{0.4, 0.2, 0.1, 0.01\}$, i.e., the same selected run-time threshold was used to evaluate the confidence of $M_0$, $M_1$ and $M_2$ (for the three-member ensemble).

    \subsection{Results}

        \subsubsection{Two-member ensembles}

\begin{figure}
	   \includegraphics[width=\linewidth]{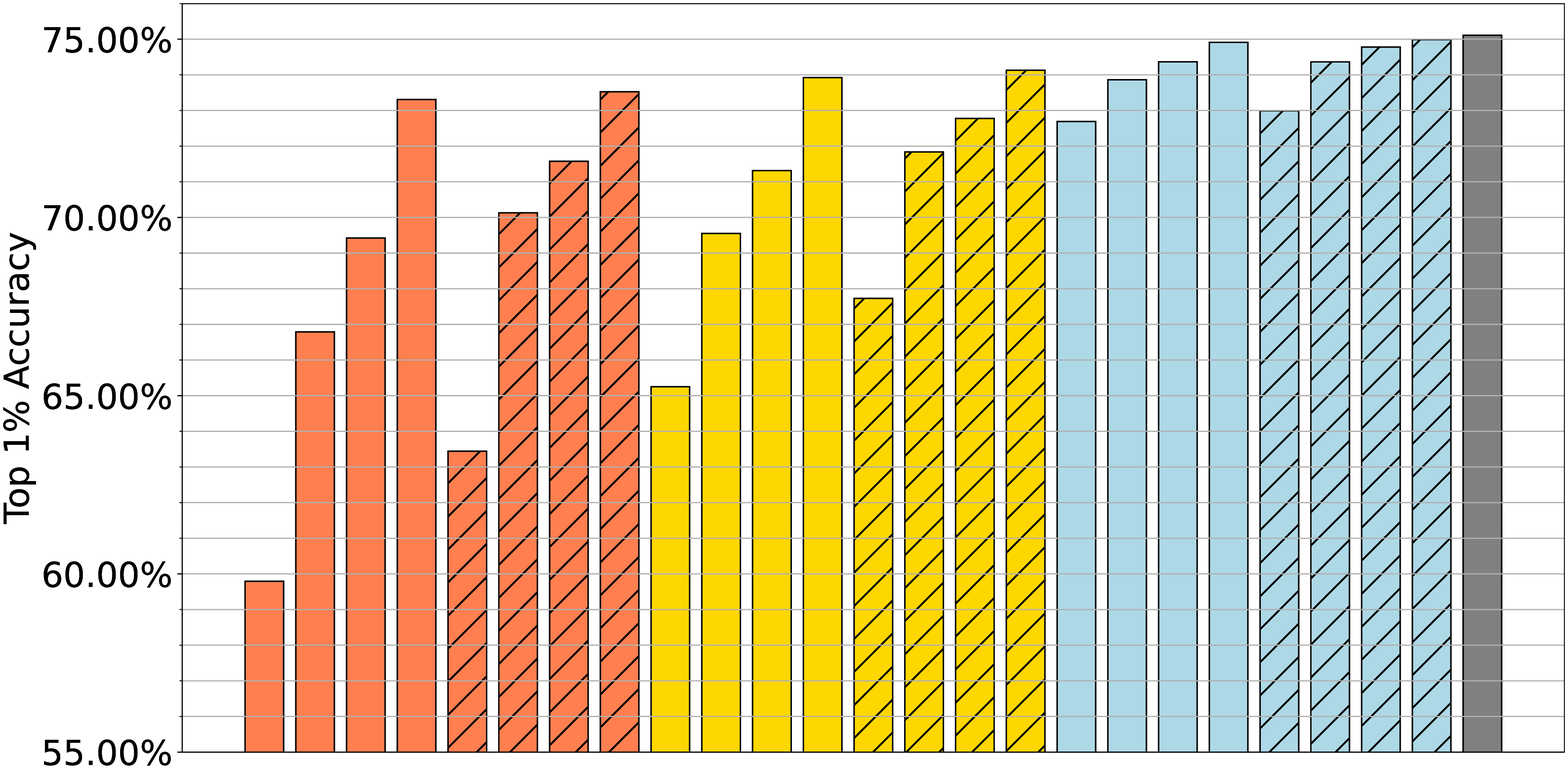}%
    \llap{%
    \raisebox{-1.3cm}{\rotatebox{90}{\parbox{3cm}{$T_{r}^{1}=0.01$}}}%
    \hspace{0.01cm}
    \raisebox{-1.3cm}{\rotatebox{90}{\parbox{3cm}{$T_{r}^{1}=0.10$}}}%
    \hspace{0.01cm}
    \raisebox{-1.3cm}{\rotatebox{90}{\parbox{3cm}{$T_{r}^{1}=0.20$}}}%
    \hspace{0.01cm}
    \raisebox{-1.3cm}{\rotatebox{90}{\parbox{3cm}{$T_{r}^{1}=0.40$}}}%
    \hspace{1.45cm} 
    \raisebox{-1.3cm}{\rotatebox{90}{\parbox{3cm}{$T_{r}^{1}=0.01$}}}%
    \hspace{0.01cm}
    \raisebox{-1.3cm}{\rotatebox{90}{\parbox{3cm}{$T_{r}^{1}=0.10$}}}%
    \hspace{0.01cm}
    \raisebox{-1.3cm}{\rotatebox{90}{\parbox{3cm}{$T_{r}^{1}=0.20$}}}%
    \hspace{0.01cm}
    \raisebox{-1.3cm}{\rotatebox{90}{\parbox{3cm}{$T_{r}^{1}=0.40$}}}%
    \hspace{1.45cm} 
    \raisebox{-1.3cm}{\rotatebox{90}{\parbox{3cm}{$T_{r}^{1}=0.01$}}}%
    \hspace{0.01cm}
    \raisebox{-1.3cm}{\rotatebox{90}{\parbox{3cm}{$T_{r}^{1}=0.10$}}}%
    \hspace{0.01cm}
    \raisebox{-1.3cm}{\rotatebox{90}{\parbox{3cm}{$T_{r}^{1}=0.20$}}}%
    \hspace{0.01cm}
    \raisebox{-1.3cm}{\rotatebox{90}{\parbox{3cm}{$T_{r}^{1}=0.40$}}}%
    \hspace{-.45cm}
    \raisebox{1cm}{\includegraphics[height=2cm]{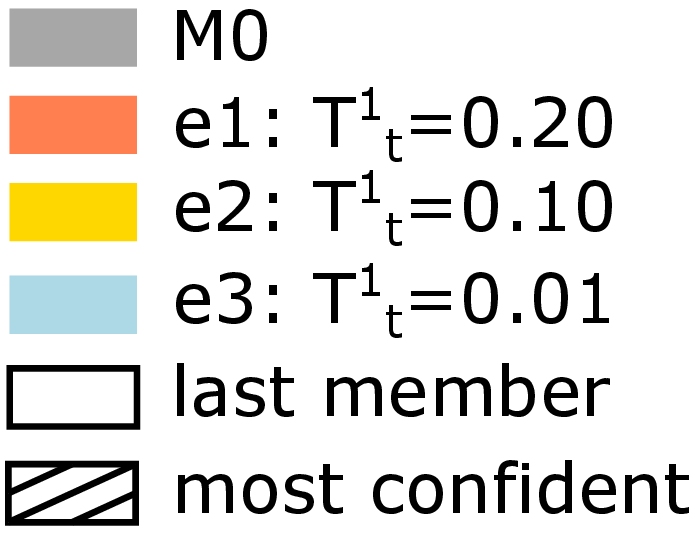}}
          }
	   \caption{Comparison of the accuracy of different two-member ensembles. Different combinations of run-time and training thresholds are compared using two different consensus mechanisms. For each training threshold (color-coded) four different run-time thresholds are evaluated (in ascending order). The two consensus mechanisms are distinguished with different hatches (clean or lines). For reference, the baseline single model $M_0$ is shown in grey too. This figure shows that both higher run-time thresholds and lower run-time thresholds result in more accurate ensembles.}
        \label{fig:acc}
	\end{figure}

        We compare different run-time and training threshold combinations to create a two-member ensemble and evaluate which constellation produces the most accurate ensemble according to the top1 accuracy metric, i.e., the correctness of the prediction with the most probability according to the ensemble.
        Figure~\ref{fig:acc} shows, for different run-time thresholds and consensus heuristics, the top1 accuracy results of the two-member ensembles created with the confidence thresholds for $M_1$ defined in Section~\ref{sec:design}.
        
        Observations to Figure~\ref{fig:acc}: 
        \begin{itemize}
            \item The first main observation is that none of these ensembles was superior to the original model $M_0$ (grey).
            \item Lower training thresholds work better.
            The ensemble $e3$ (blue) with training threshold $T_{r}^{L=1}=0.01$ achieved higher accuracies.
            A lower training threshold means more data samples are used for the training of the next members.
            \item Higher run-time thresholds work better.
            A high run-time threshold means that more predictions are taken from the initial ensemble member.
            \item Most confident consensus was better than last evaluated consensus.
        \end{itemize}
        
        \paragraph{Inspection of confidence scores}

        To better understand why the combination of low training threshold and high run-time threshold produced the best (but insufficient) results, we investigate the statistics of the correctness and confidence in the classification of the individual models.

        We compute the histogram on the entire training dataset of the uncertainty score and of the probability score.
        The histogram shows the correct and incorrect classifications in different colors (green and red respectively).

        \begin{figure}[htp]
        \subcaptionbox{\centering$M_1$ trained with $T_{t}^{1}=0.20$\label{fig:confM1C2}}{\includegraphics[width=0.25\linewidth]{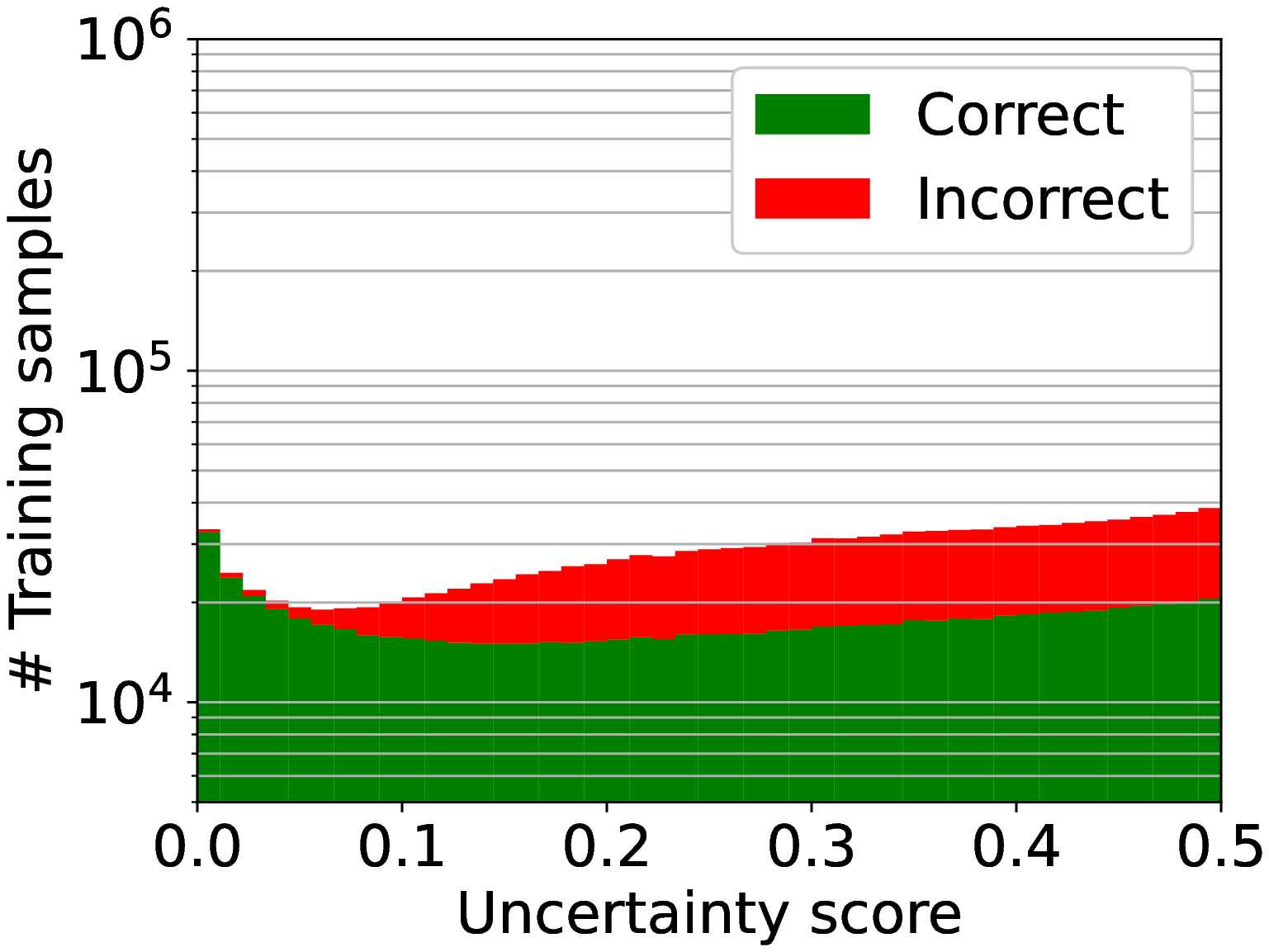}}\hspace{0em}%
        \subcaptionbox{\centering$M_1$ trained with $T_{t}^{1}=0.10$\label{fig:confM1C1}}{\includegraphics[width=0.25\linewidth]{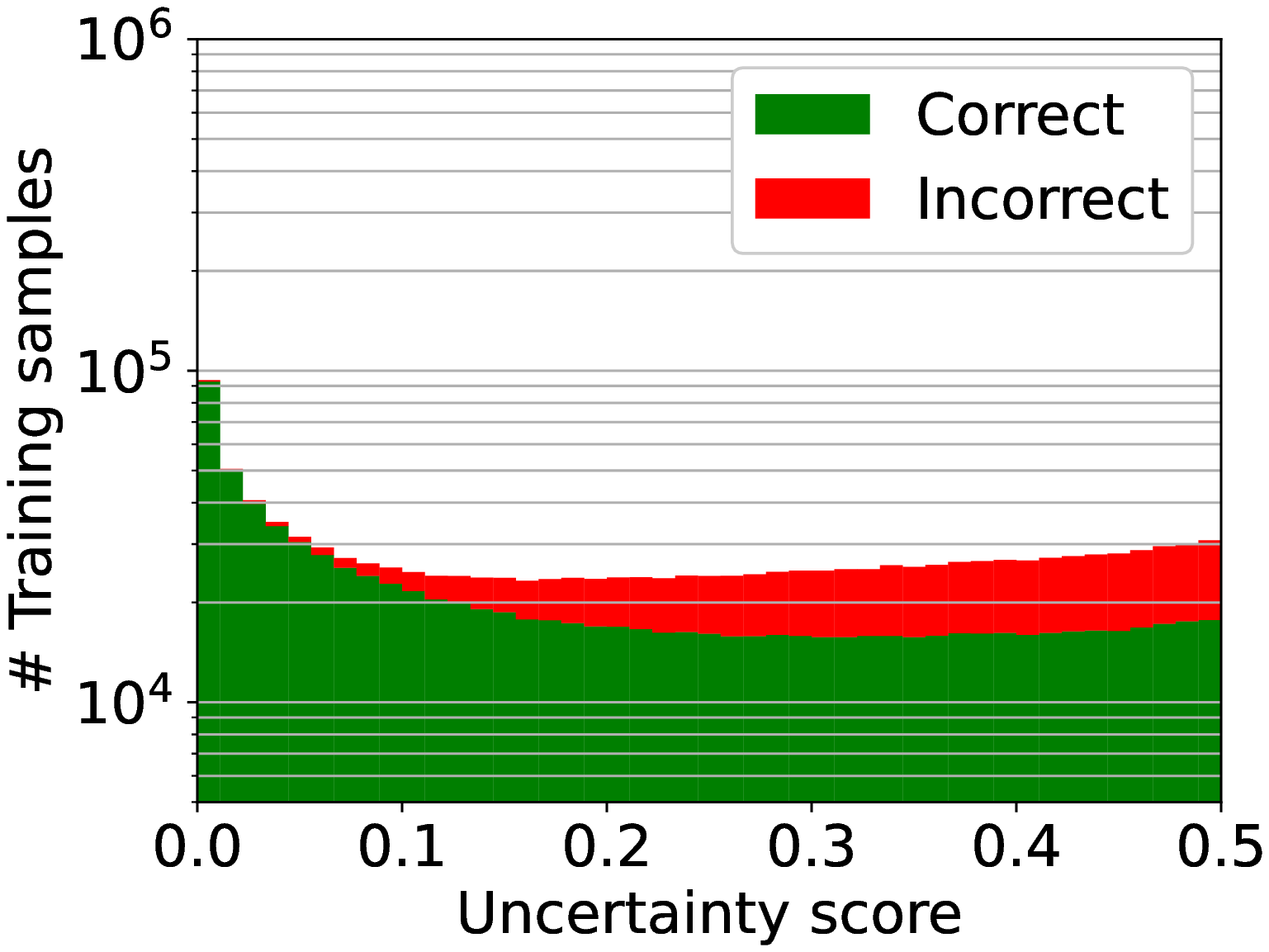}}\hspace{0em}%
        \subcaptionbox{\centering$M_1$ trained with $T_{t}^{1}=0.01$\label{fig:confM1C01}}{\includegraphics[width=0.25\linewidth]{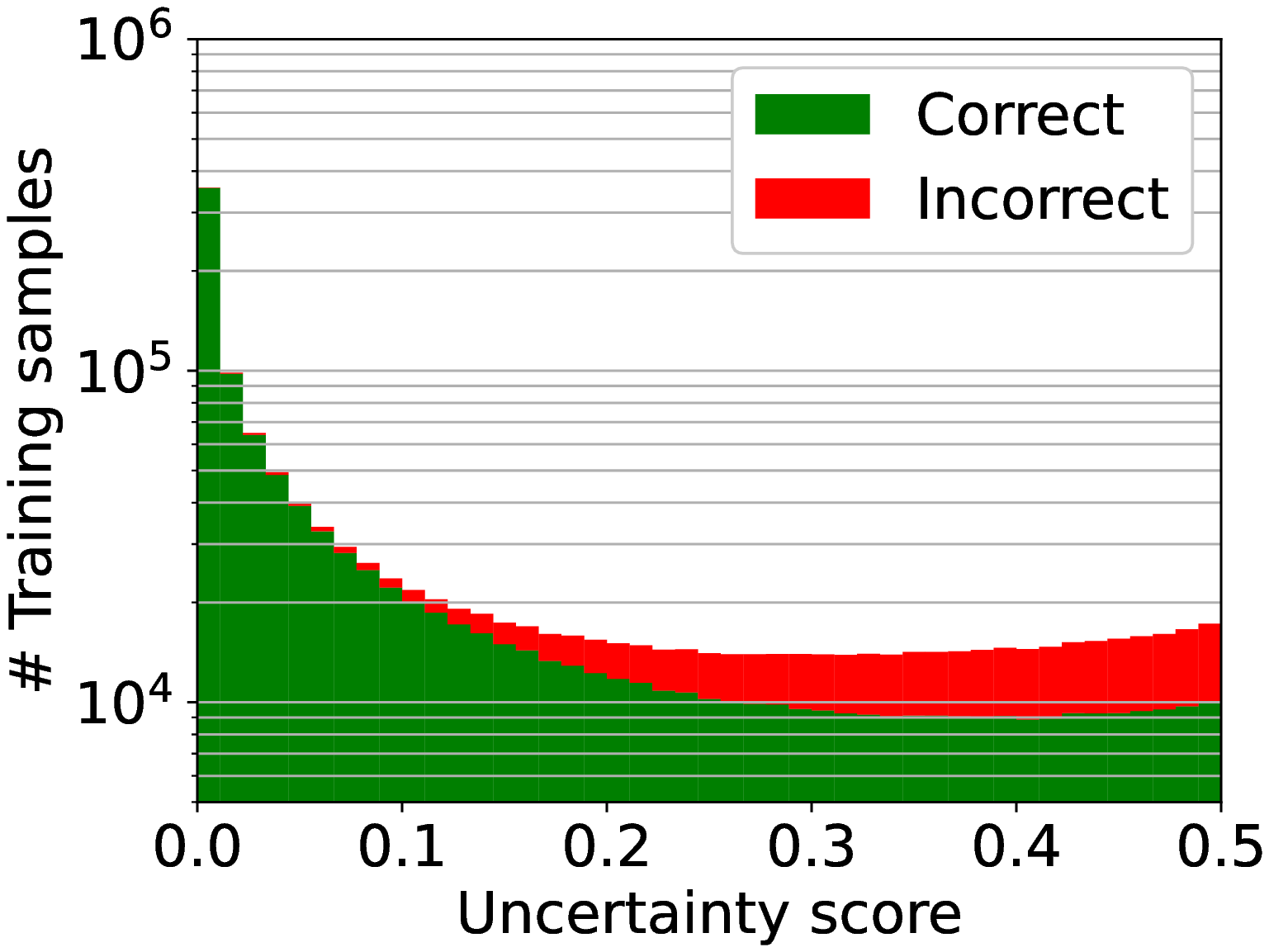}}\hspace{0em}%
        \subcaptionbox{$M_0$\label{fig:confM0}}{\includegraphics[width=0.25\linewidth]{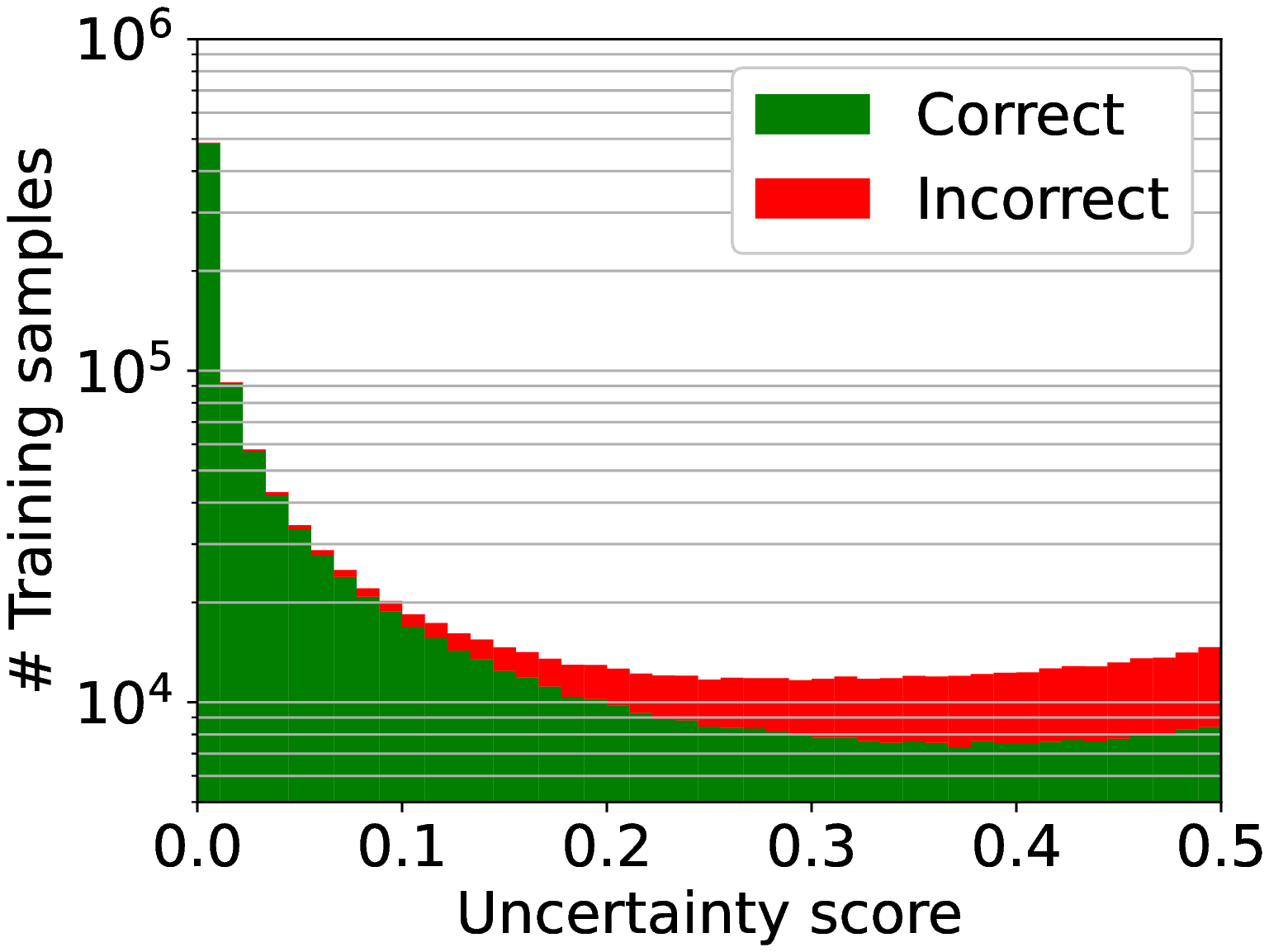}}
        \caption{Confidence score histogram of a single ensemble member trained with a given confidence threshold.
        $M_0$ is the first model trained with the entire dataset (no threshold). $M_1$ is the next ensemble member trained with a subset of the dataset where $M_0$'s unconfidence score was above the respective threshold.} \label{fig:conf}
        \end{figure}
        
\begin{figure}[htp]
        \subcaptionbox{\centering$M_1$ trained with $T_{t}^{1}=0.20$\label{fig:softM1C2}}{\includegraphics[width=0.25\linewidth]{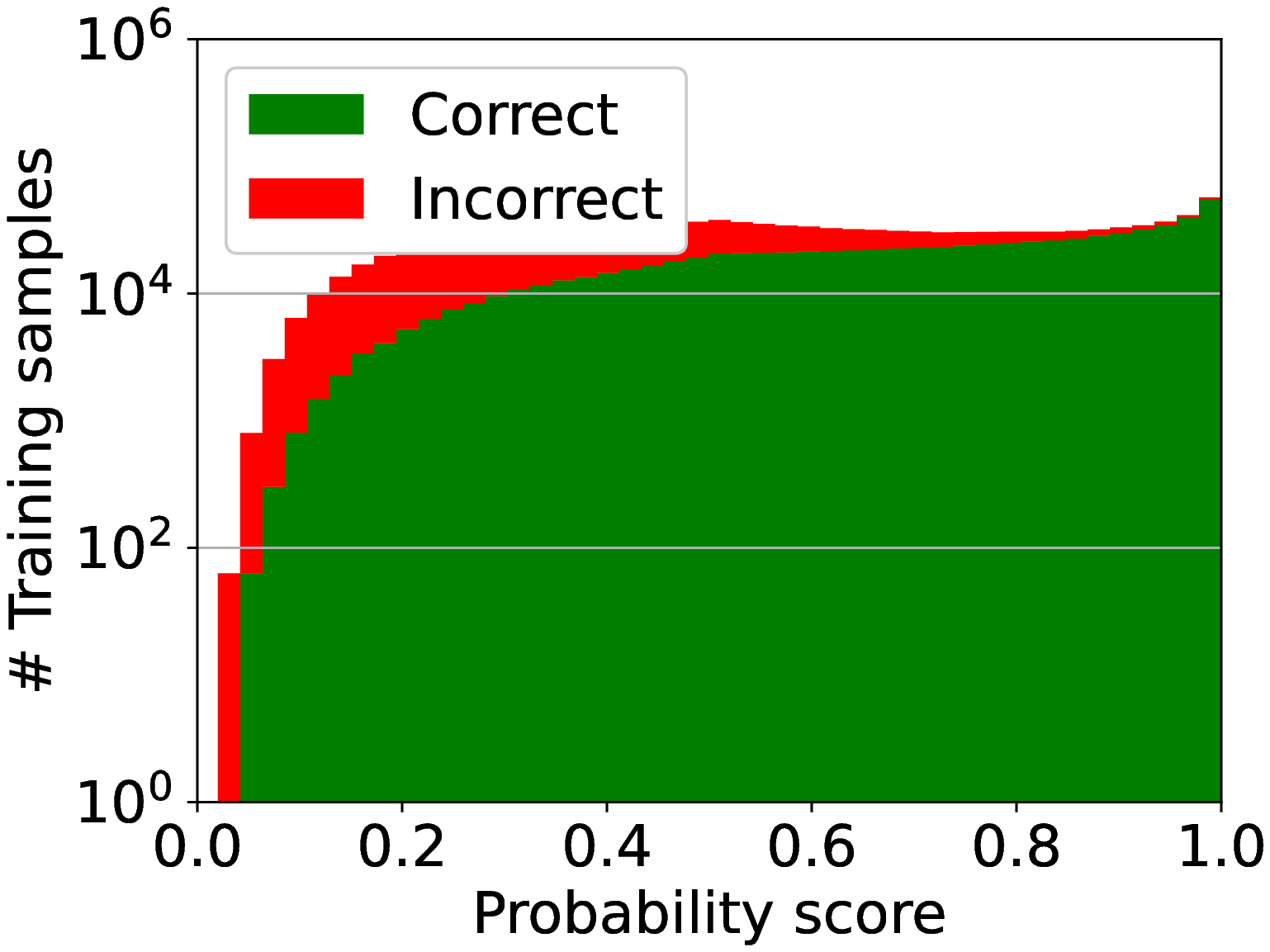}}\hspace{0em}%
        \subcaptionbox{\centering$M_1$ trained with $T_{t}^{1}=0.10$\label{fig:softM1C1}}{\includegraphics[width=0.25\linewidth]{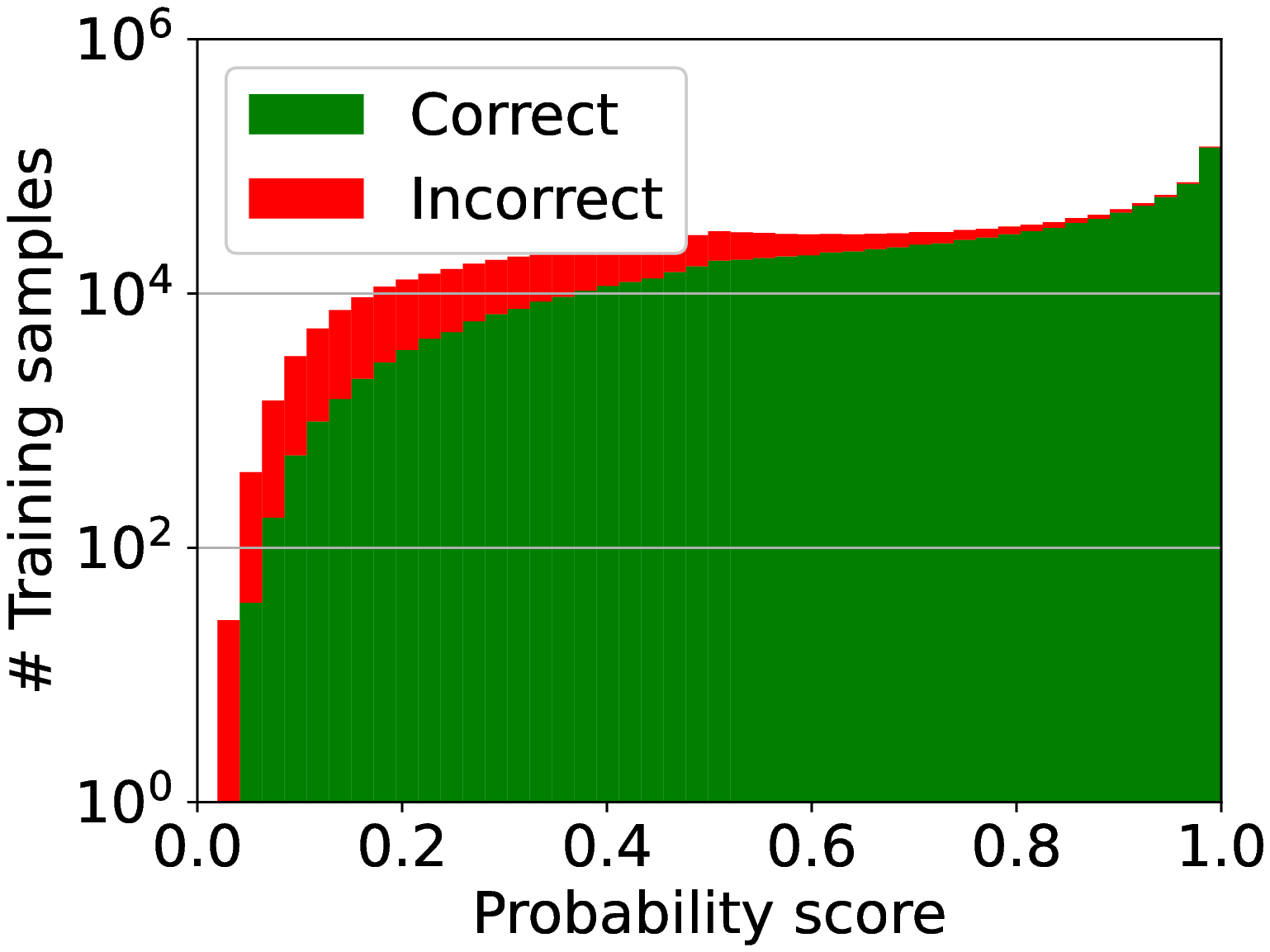}}\hspace{0em}%
        \subcaptionbox{\centering$M_1$ trained with $T_{t}^{1}=0.01$\label{fig:softM1C01}}{\includegraphics[width=0.25\linewidth]{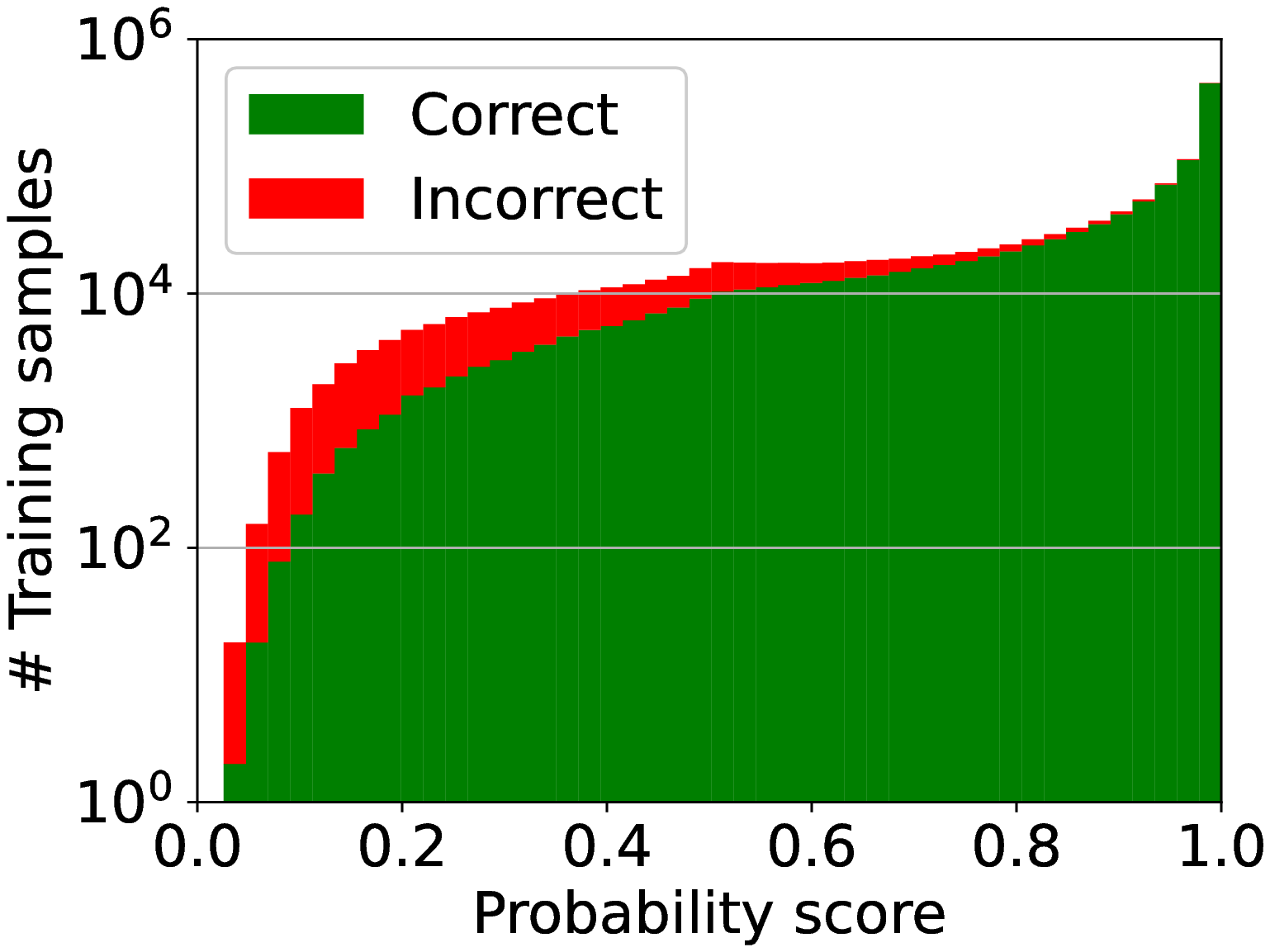}}\hspace{0em}%
        \subcaptionbox{$M_0$\label{fig:softM0}}{\includegraphics[width=0.25\linewidth]{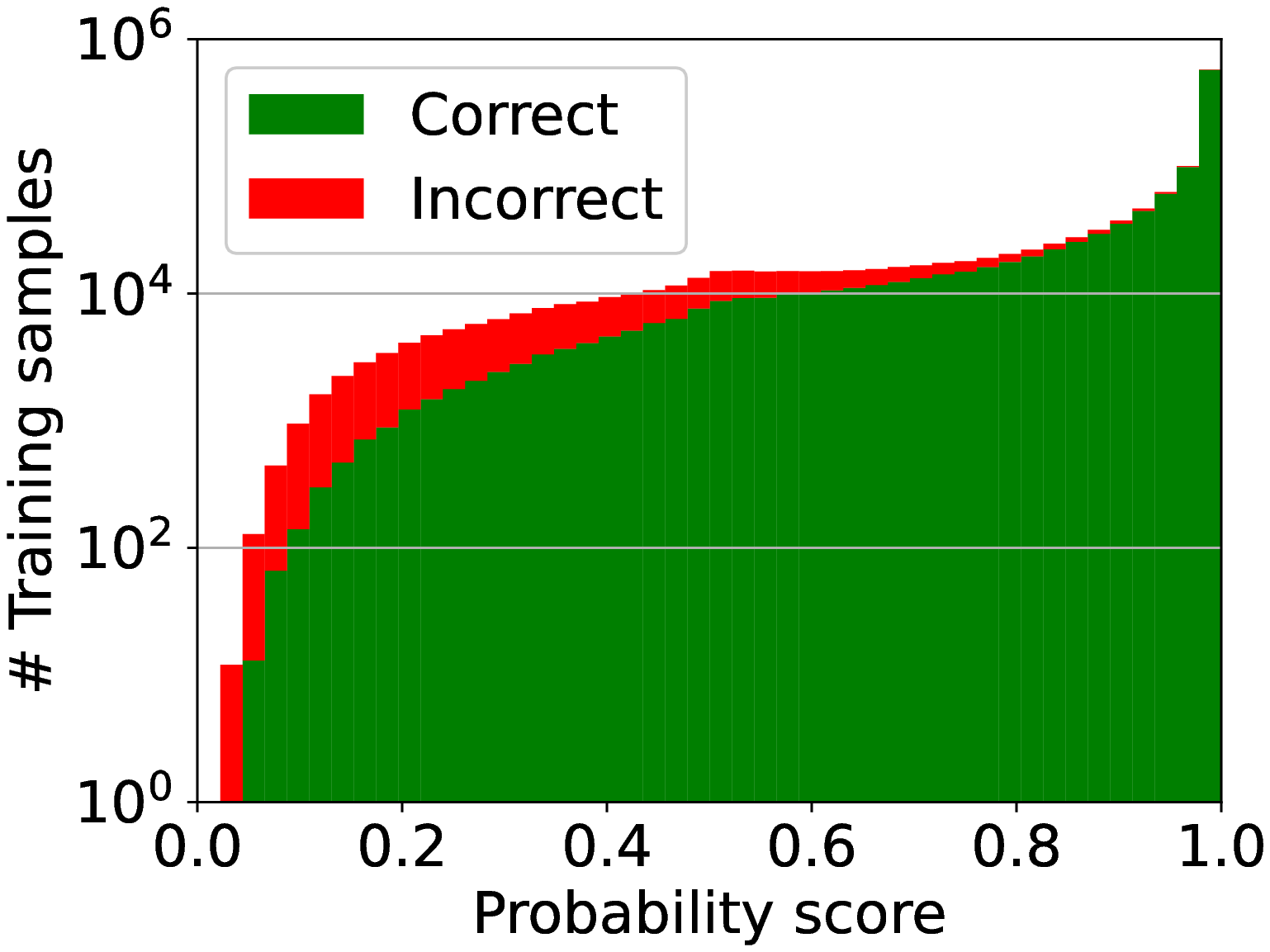}}
        \caption{Probability score (output of softmax layer) histogram of a single ensemble member trained with a given confidence threshold} \label{fig:softmax}
        \end{figure}

        Figure~\ref{fig:conf} shows the confidence scores of $M_1$ (for each training threshold) and $M_0$.
        Likewise, Figure~\ref{fig:softmax} shows the probability scores of $M_1$ (for each training threshold) and $M_0$.
        
        Observations to Figure~\ref{fig:conf} and~\ref{fig:softmax}:
        \begin{itemize}
            \item The number of confident predictions for correct classifications is negatively correlated with the training threshold.
            The peak of high number of correct predictions at low uncertainty scores diminishes as the training threshold is increased.

            \item  The number of confident predictions of incorrect classifications are positively correlated with increasing training threshold.
            The number of incorrect predictions at low uncertainty scores increases as the training threshold is increased.

        \end{itemize}

        \subsubsection{Three member ensemble}

        Based on our observations from a two-member ensemble, we created a third member with a modified heuristic
        as specified in Section~\ref{sec:propExt}: instead of using ever decreasing datasets for the training of the subsequent ensemble members, we use the proposed heuristic, see Equation~\ref{eq:proptrainingset},  to always use the original dataset as a reference and from there select a subset of samples according to the confidence values of the previous member on the \textit{whole} dataset.

        \begin{figure}
	   \includegraphics[width=\linewidth]{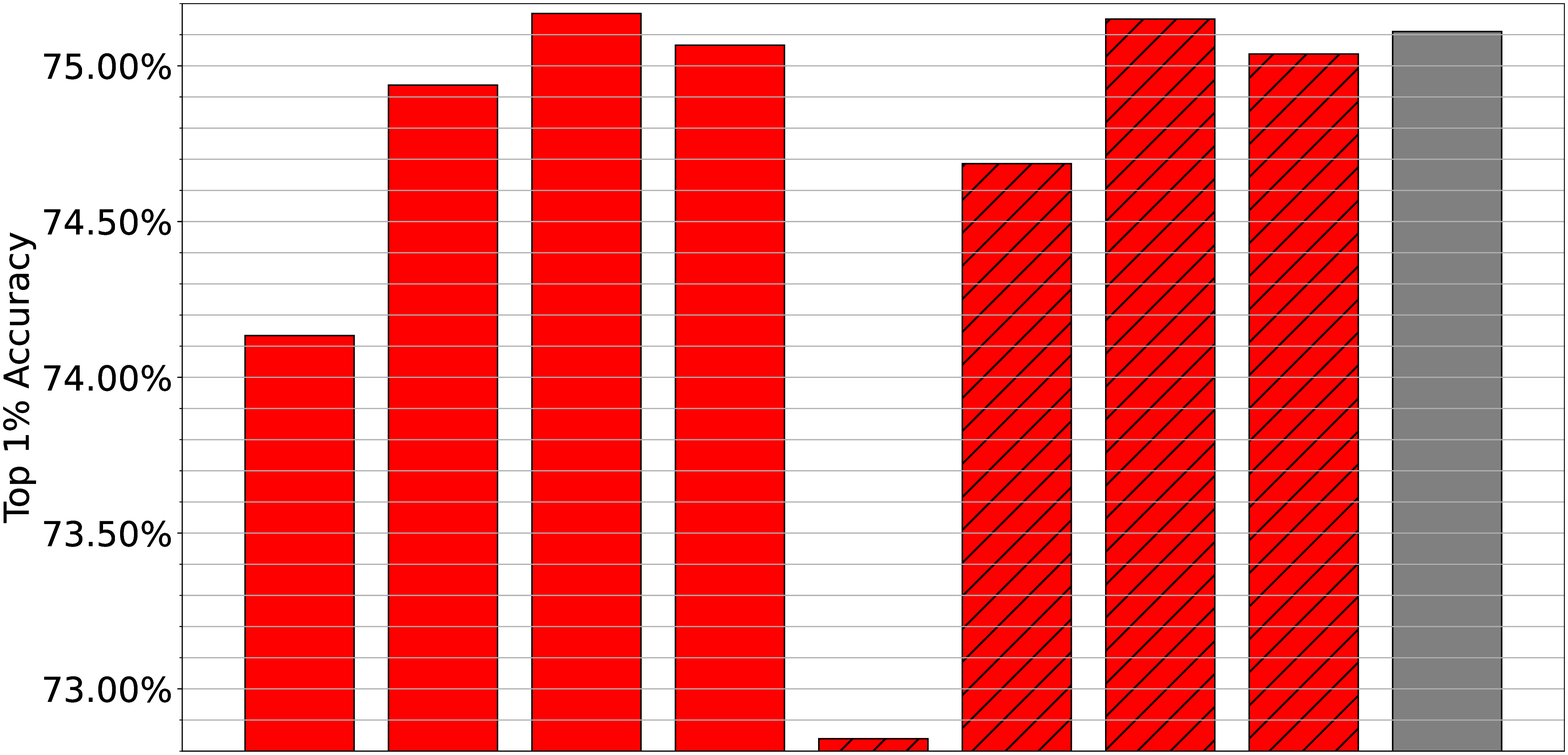}%
    \llap{%
    \raisebox{-1.5cm}{\rotatebox{90}{\parbox{3cm}{$T_{r}^{2}=0.01$}}}%
    \hspace{0.70cm}
    \raisebox{-1.5cm}{\rotatebox{90}{\parbox{3cm}{$T_{r}^{2}=0.10$}}}%
    \hspace{0.70cm}
    \raisebox{-1.5cm}{\rotatebox{90}{\parbox{3cm}{$T_{r}^{2}=0.20$}}}%
    \hspace{0.70cm}
    \raisebox{-1.5cm}{\rotatebox{90}{\parbox{3cm}{$T_{r}^{2}=0.40$}}}%
    \hspace{0.70cm} 
    \raisebox{-1.5cm}{\rotatebox{90}{\parbox{3cm}{$T_{r}^{2}=0.01$}}}%
    \hspace{0.70cm}
    \raisebox{-1.5cm}{\rotatebox{90}{\parbox{3cm}{$T_{r}^{2}=0.10$}}}%
    \hspace{0.70cm}
    \raisebox{-1.5cm}{\rotatebox{90}{\parbox{3cm}{$T_{r}^{2}=0.20$}}}%
    \hspace{0.70cm}
    \raisebox{-1.5cm}{\rotatebox{90}{\parbox{3cm}{$T_{r}^{2}=0.40$}}}%
    \hspace{-0.85cm}
    \raisebox{1cm}{\includegraphics[height=1.1cm]{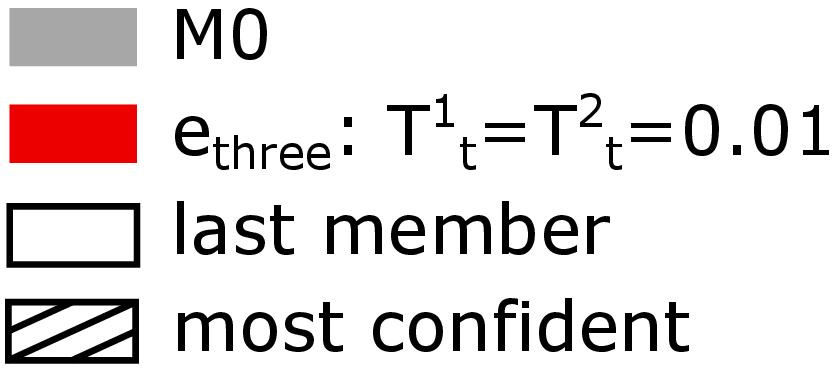}
    }
          }
	   \caption{Accuracy of three-member ensembles. The ensembles created with run-time threshold values of 0.20 surpass the accuracy of the original model.}
        \label{fig:accThreeMem}
	\end{figure}
        
        Figure~\ref{fig:accThreeMem} shows the top1 accuracy, for different run-time thresholds and consensus heuristics, of the three-member ensemble created using a training threshold of $0.01$ as the empirical results of the two-member ensemble indicated better results.

        Observations to Figure~\ref{fig:accThreeMem}:
        \begin{itemize}
            \item The three member ensemble was capable to surpass the single member ensemble $M_0$ by 0.10\%
            \item In this case, the highest run-time threshold was not the best, here $T_{r}^{L=2}=0.2$ was better.
            This indicates that using the subsequent ensemble members is useful and a sweet point is not at the extremes, i.e., a balance must be found on how much of the uncertain inference samples should be classified by the earlier models.
            \end{itemize}

        \subsubsection{Calibration}

\begin{figure}
	   \includegraphics[width=\linewidth]{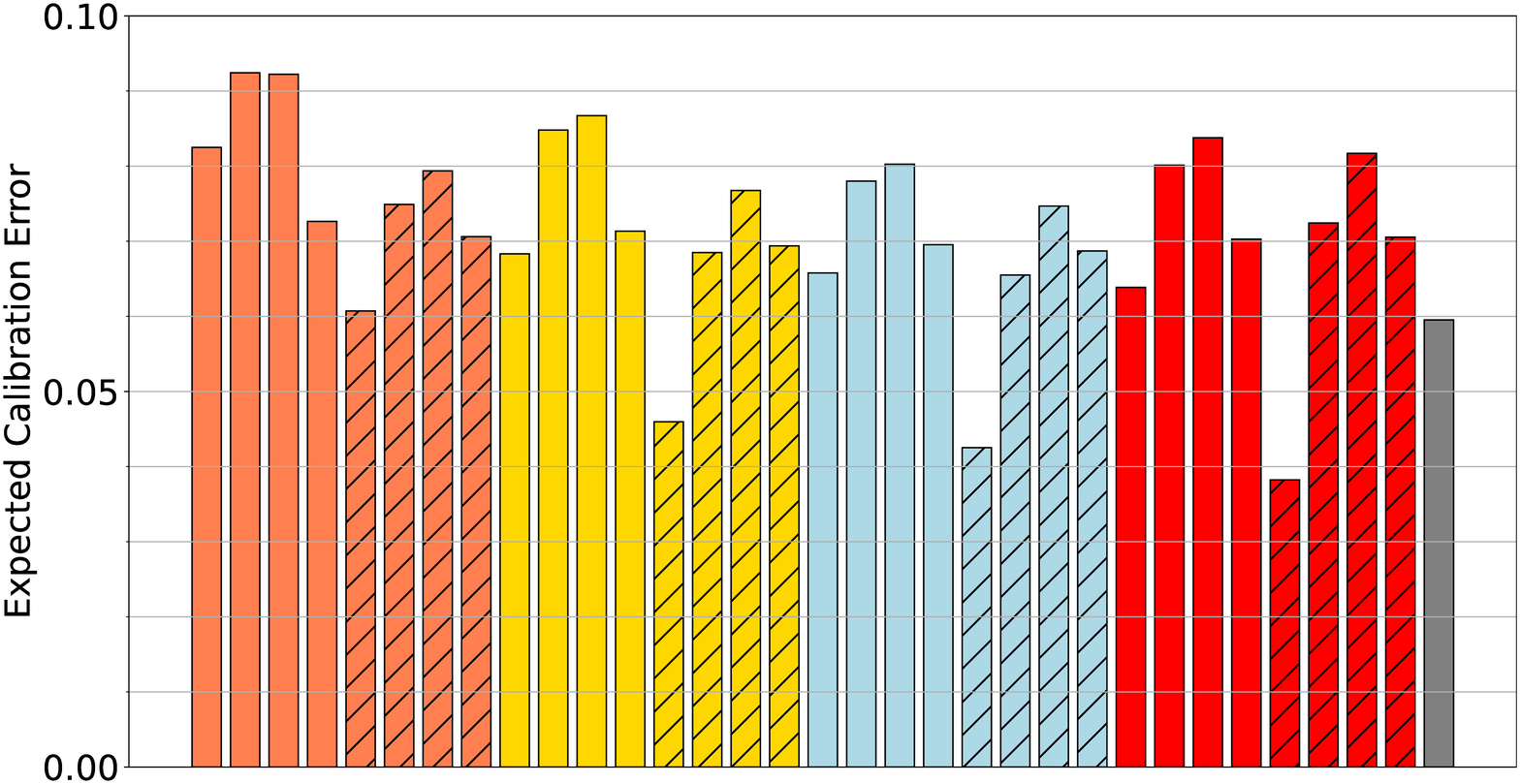}%
    \llap{%
    \raisebox{1cm}{\includegraphics[height=2.5cm]{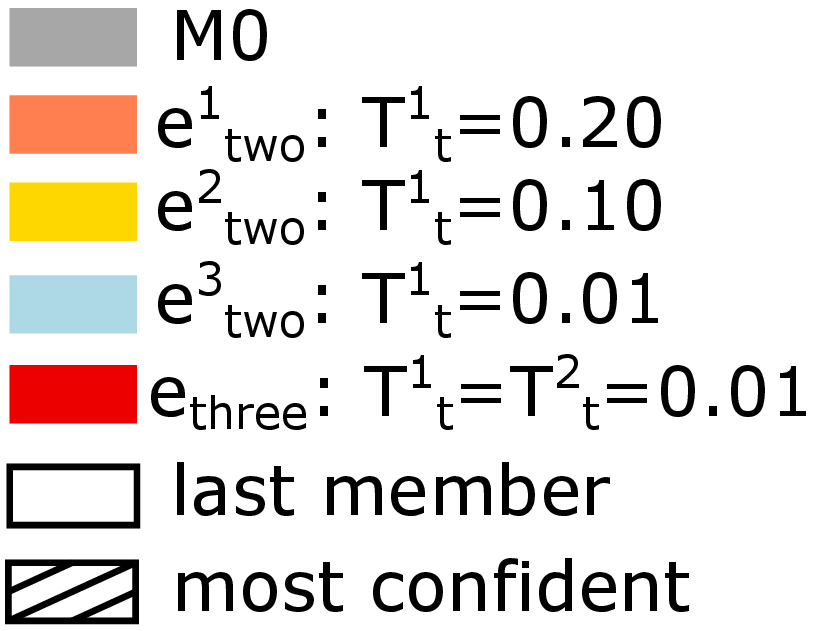}}%
    \hspace{-3.0cm}
    \raisebox{-1.3cm}{\rotatebox{90}{\parbox{3cm}{$T_{r}^{1}=0.01$}}}%
    \raisebox{-1.3cm}{\rotatebox{90}{\parbox{3cm}{$T_{r}^{1}=0.10$}}}%
    \raisebox{-1.3cm}{\rotatebox{90}{\parbox{3cm}{$T_{r}^{1}=0.20$}}}%
    \raisebox{-1.3cm}{\rotatebox{90}{\parbox{3cm}{$T_{r}^{1}=0.40$}}}%
    \hspace{1.05cm} 
    \raisebox{-1.3cm}{\rotatebox{90}{\parbox{3cm}{$T_{r}^{1}=0.01$}}}%
    \raisebox{-1.3cm}{\rotatebox{90}{\parbox{3cm}{$T_{r}^{1}=0.10$}}}%
    \raisebox{-1.3cm}{\rotatebox{90}{\parbox{3cm}{$T_{r}^{1}=0.20$}}}%
    \raisebox{-1.3cm}{\rotatebox{90}{\parbox{3cm}{$T_{r}^{1}=0.40$}}}%
    \hspace{1.0cm} 
    \raisebox{-1.3cm}{\rotatebox{90}{\parbox{3cm}{$T_{r}^{1}=0.01$}}}%
    \raisebox{-1.3cm}{\rotatebox{90}{\parbox{3cm}{$T_{r}^{1}=0.10$}}}%
    \raisebox{-1.3cm}{\rotatebox{90}{\parbox{3cm}{$T_{r}^{1}=0.20$}}}%
    \raisebox{-1.3cm}{\rotatebox{90}{\parbox{3cm}{$T_{r}^{1}=0.40$}}}%
    \hspace{1.1cm} 
    \raisebox{-1.3cm}{\rotatebox{90}{\parbox{3cm}{$T_{r}^{2}=0.01$}}}%
    \raisebox{-1.3cm}{\rotatebox{90}{\parbox{3cm}{$T_{r}^{2}=0.10$}}}%
    \raisebox{-1.3cm}{\rotatebox{90}{\parbox{3cm}{$T_{r}^{2}=0.20$}}}%
    \raisebox{-1.3cm}{\rotatebox{90}{\parbox{3cm}{$T_{r}^{2}=0.40$}}}%
    \hspace{1.90cm}
    }
    \caption{Expected Calibration Error of all ensembles. The Ensembles with a lower run-time training threshold had a lower expected calibration error than a single model (grey).}
        \label{fig:ece}
	\end{figure}

        Finally we display the Expected Calibration Error (ECE)  metric~\cite{NaeiniCH15} $ECE=\sum_{i=1}^{K}{P(i)\cdot\left|o_{i} - e_{i} \right|}$, where $o_{i}$ is the number of correct instances in bin $i$, $e_{i}$ is the mean of the probability scores in bin $i$, and $P(i)$ is the probability of all instances that fall into bin $i$.
        It provides an estimation of how correlated are the model probability scores to the correctness of the classification.
        A lower ECE score is better.
        Figure~\ref{fig:ece} shows the ECE metric for all ensembles on different run-time thresholds and consensus heuristics.

        Observations to Figure~\ref{fig:ece}:
        \begin{itemize}
            \item We observe that lower training thresholds tend to produce better calibrated ensembles.
            \item We observe that low run-time thresholds tend to produce better calibrated ensembles even though their accuracy is not the best.
        \end{itemize}

\section{Discussion and conclusions}
\label{sec:discConc}

    We now discuss the obtained results by lessons learned, i.e., what consensus heuristic is more effective, and what impact the value of confidence thresholds have in the resulting accuracy of the ensemble.

    \subsection{Best consensus heuristic for image classification}

    It was consistently confirmed empirically that the most confident heuristic is better than selecting the prediction of the last ensemble, which is not surprising.
    This suggests that the believe that the later ensemble members would be on average more adequate to classify difficult instances for which they were specifically trained for is not true.

    \subsection{Best training threshold for image classification}

    The fact that a lower training threshold is better in all our experiments means that more data (not only uncertain samples) fed to all models is better.
    This leads us to speculate that feeding more data could be:
    \begin{itemize}
        \item Mitigating problems in the training data set, i.e, less confident predictions may be due to miss-labels, and if only this samples are used for training, the result can be worse.
        Miss-labeling is a known challenge of supervised learning in big data due to the cost of curating an enormous corpus of data, and ImageNet is known to have miss-label problems in about 3\% of the training data~\cite{northcutt2021pervasive}.
        \item Small training data sets, even though correct, are not enough to create a robust classifier for complex multi-class. In the original approach of Conf-Ensemble the problem of having too little data for the subsequent ensemble members should be mitigated by obtaining more training data.
        In this study however, we evaluate the task of image classification with an assumed limited amount of training data with ImageNet as the typical reference baseline, and thus propose an adaptation of Conf-Ensembles to mitigate this problem.
        \end{itemize}
    We suspect that for this complex image classification task, a lot of data is needed to discern complex cases and filter labeling noise.

    \subsection{Best run-time threshold for image classification}

    A higher run-time threshold tends to favor the use of earlier models that were trained with more data.
    However, as evident from our findings (Figure~\ref{fig:accThreeMem})  there exists a \textit{sweet point} of a threshold value where the accuracy is maximized by actually using the other members of the ensemble.
    
    \subsection{Conclusions}

    Conf-Ensemble is a principled approach, which has been demonstrated useful, in binary classification.
    For the multi-label image classification vision task, key to machine learning based perception in safety-critical systems, it seems that the amount of training data for each member must be kept high to obtain a good ensemble.
    The hyper-parameters for training and run-time inference to deem a prediction confident or uncertain are key in the final performance of the ensemble, and their optimal values must be empirically determined.
    We have seen that an ensemble of three members, where the training data sets of each member were suitably selected subsets of the training set accounting for the confident classification of samples by the previous member of the ensemble, is capable to provide more accurate classification than a single model, albeit the gain is small.
    This result suggest Conf-Ensemble has the potential to be useful with complex image classification tasks. 
    In this paper we report preliminary results with Conf-Ensemble used for image classification.
    Although encouraging, result are not conclusive and further work is needed. We briefly summarise direction for future research.
    \begin{itemize}
        \item Tuning the members of Conf-Ensemble using the  training and classification thresholds must be done systematically.
        We identified that these two ensemble parameters affect significantly ensembles performance, but systematic study of how these two parameters affect the ensemble performance, possibly how they are interrelated is yet to be conducted.
        \item Applying Conf-Ensemble in its pure form~\cite{confEns} leads to a reduction of the training set for the "later" layers of the ensemble.
        A small training dataset limits accuracy of the ML classifier. In this study we adopted an ad-hoc modification which led to training of "later" layers using samples which these "later" layers are unlikely to see at run-time as a significant proportion may be classified with high confidence by the previous layers.
        As a result of the adopted modification in training the later members we make them bias, i.e., trained on a dataset which is systematically different from what these layers will be asked to classify at run time.
        This apparent contradiction between "focused" training and training on sufficiently large training dataset is yet to be addressed comprehensively.   
    \end{itemize}
\subsubsection{Acknowledgements}

This work was partially funded by the Federal Ministry for Economic Affairs and Climate Action of Germany, as part of the research project SafeWahr (Grant Number: 19A21026C)

\bibliographystyle{splncs04}
\bibliography{references}

\newpage

\end{document}